\DeclareSymbolFont{LMcal}{OMS}{lmsy}{m}{n}
\DeclareSymbolFontAlphabet{\mathcalLM}{LMcal}
\documentclass[12pt]{article}
\usepackage{newtxtext,newtxmath}
\usepackage{indentfirst}
\usepackage{graphicx}
\usepackage{caption}
\usepackage{tabularx}
\usepackage{array}
\usepackage{makecell}
\usepackage{float}
\usepackage{lastpage}
\usepackage{pifont}
\usepackage{tabularx}
\usepackage{hyperref}
\usepackage{overpic}
\usepackage{needspace}
\usepackage{calrsfs}
\usepackage{mathrsfs}
\usepackage{amsmath}
\usepackage{makecell}
\usepackage{tabularx}
\usepackage{array}
\usepackage[letterpaper, left=0.8in, right=0.8in, top=1in, bottom=1in]{geometry}
\renewenvironment{abstract}
	{\quotation}
	{\endquotation}

\date{}

\makeatletter
\renewcommand{\fnum@figure}{\textbf{Figure \thefigure}}
\renewcommand{\fnum@table}{\textbf{Table \thetable}}
\makeatother

\usepackage{scicite}
\usepackage{url}

\def\scititle{
Switchable-Polarity Electropermanent Magnet: Reconfigurable Magnetic Fields for Scalable Fluidic Control
}
\title{\bfseries \boldmath \scititle}

\author{
	Bingchao~Wang$^{1}$,
	Jonah~Mack$^{1}$,
	Francesco~Giorgio-Serchi$^{1}$,
	Adam~A.~Stokes$^{1\ast}$\and
	\small$^{1}$Soft Systems Group, The School of Engineering, The University of Edinburgh, EH9 3FF Edinburgh, UK.\and
	\small$^\ast$Corresponding author. Email: adam.stokes@ed.ac.uk
}

\begin{document}
\maketitle

\begin{abstract} \bfseries \boldmath
Scalable control of pneumatic and fluidic networks remains fundamentally constrained by architectures that require continuous power input, dense external control hardware, and fixed routing topologies. Current valve arrays rely on such continuous actuation and mechanically fixed routing, imposing substantial thermal and architectural overhead. Here, we introduce the Switchable-polarity Electropermanent Magnet (S-EPM), a bistable magnetic architecture that spatially redistributes its external magnetic field through transient electrical excitation. By reconfiguring internal flux pathways within a composite magnet assembly, the S-EPM exchanges reinforced and weakened magnetic-field regions while maintaining bistable, zero-power state retention. We integrate this architecture into a compact pinch-valve to robustly control pneumatic and liquid media. This state-encoded magnetic control enables programmable fluidic networks, including decoders, hierarchical distribution modules, and a nonvolatile six-port routing array. By embedding functionality in persistent magnetic states rather than continuous power or static plumbing, this work establishes a scalable foundation for programmable fluidics and self-driving laboratories. 
\end{abstract}

\noindent
Advanced applications in automated chemical processing, self-driving laboratories, biomedical diagnostics, and fluid-driven soft robotics require efficient fluid routing and multi-channel switching \cite{bios12111023,D4LC00423J,10.3389/frobt.2021.720702,D3DD00223C}. In chemical automation, dynamic routing enables diverse reaction sequences and parallel sample processing \cite{Schuster2020, Kerk2021}, while in soft robotics, multi-channel actuation coordinates motion across multiple degrees of freedom \cite{10196254, 9785890, VANLAAKE20222898, Shin2025, 9732183}.Predominant architectures that rely upon independent pneumatic lines and external solenoid valves are bulky, thermally unstable, and difficult to scale, compromising reagent integrity and limiting robotic maneuverability \cite{Hoang2021,biomimetics3030016,Sesen2021}. Consequently, compact and programmable valve systems must efficiently switch multi-channel pneumatic and fluidic media \cite{9736571,9699012,bios12121160}.

Current fluid routing systems typically rely on mechanically actuated components, such as rotary multiport valves \cite{Oh_2006, Hinshaw2011}. Although reliable, these devices suffer from inherent limitations (mechanical complexity, large form factors, and high cost)that hinder integration into compact or portable platforms \cite{pederson2013fluid}. 3D-printed microfluidic valves offer increased design flexibility and reduced manufacturing cost \cite{C6LC00565A, mi12101247}, but they often suffer from poor sealing, limited durability under high pressure, and leakage during switching \cite{mi12101247, Compera2021}. Fluidic logic valves eliminate mechanical rotation \cite{doi:10.1126/scirobotics.aaw5496, doi:10.1126/scirobotics.adh4060, 8722755, doi:10.1073/pnas.2205922119,Gepner2025FlexPrinter}, yet these systems require continuous external control signals and lack inherent programmability for autonomous operation \cite{9044752, 9729455}.

Researchers have explored alternative actuation mechanisms using dielectric or electrorheological (ER) fluids. For instance, soft valves driven by dielectric elastomer actuators (DEAs) use Maxwell stress or electrostatic compression to modulate fluid channels \cite{doi:10.1073/pnas.2103198118, Giousouf_2013}. 
Similarly, giant electrorheological (GER) fluids or liquid metal electrodes regulate flow \cite{HUANG2022113905, Zatopa2018}. These approaches require hazardous high-voltage inputs (up to 5 kV) and involve complex fabrication processes. Magnetically actuated valves offer a lower-voltage alternative, modulating flow with magnetorheological elastomers (MREs) or fluids (MRFs) \cite{https://doi.org/10.1002/aisy.202200238, 9516875}.
Despite their potential, these magnetic systems suffer from significant thermal drift and continuous power requirements. These flaws compromise the stability of temperature-sensitive fluids. Electropermanent Magnet (EPM)-based valves offer a promising alternative by providing bistable magnetic control without continuous power consumption. Early designs used EPMs to modulate magnetorheological (MR) fluids \cite{Leps_2020, mcdonald_modulation_2022}, yet their reliance on magnetic working fluids limits their use in diverse chemical and biological processes. To broaden applicability, EPMs physically block fluid channels by moving ferrous elements \cite{marchese_soft_2011} or pinching flexible tubing \cite{Moran2024, gholizadeh_electronically_2019}. Such ``pinching'' mechanisms suit biochemical applications well, as they ensure fluid isolation and prevent contamination from metallic components \cite{Oh_2006}. Despite these benefits, existing EPM-based valves require complex fabrication. Designers typically optimize them for either pneumatic or liquid media, limiting their versatility. Furthermore, the inherent ``on-off'' nature of standard EPMs necessitates auxiliary mechanical components for bidirectional movement, hindering system integration and responsiveness\cite{Knaian2010}.
\begin{figure}%[tbhp]
\vspace{-1cm}
\centering
\includegraphics[width=0.95\linewidth]{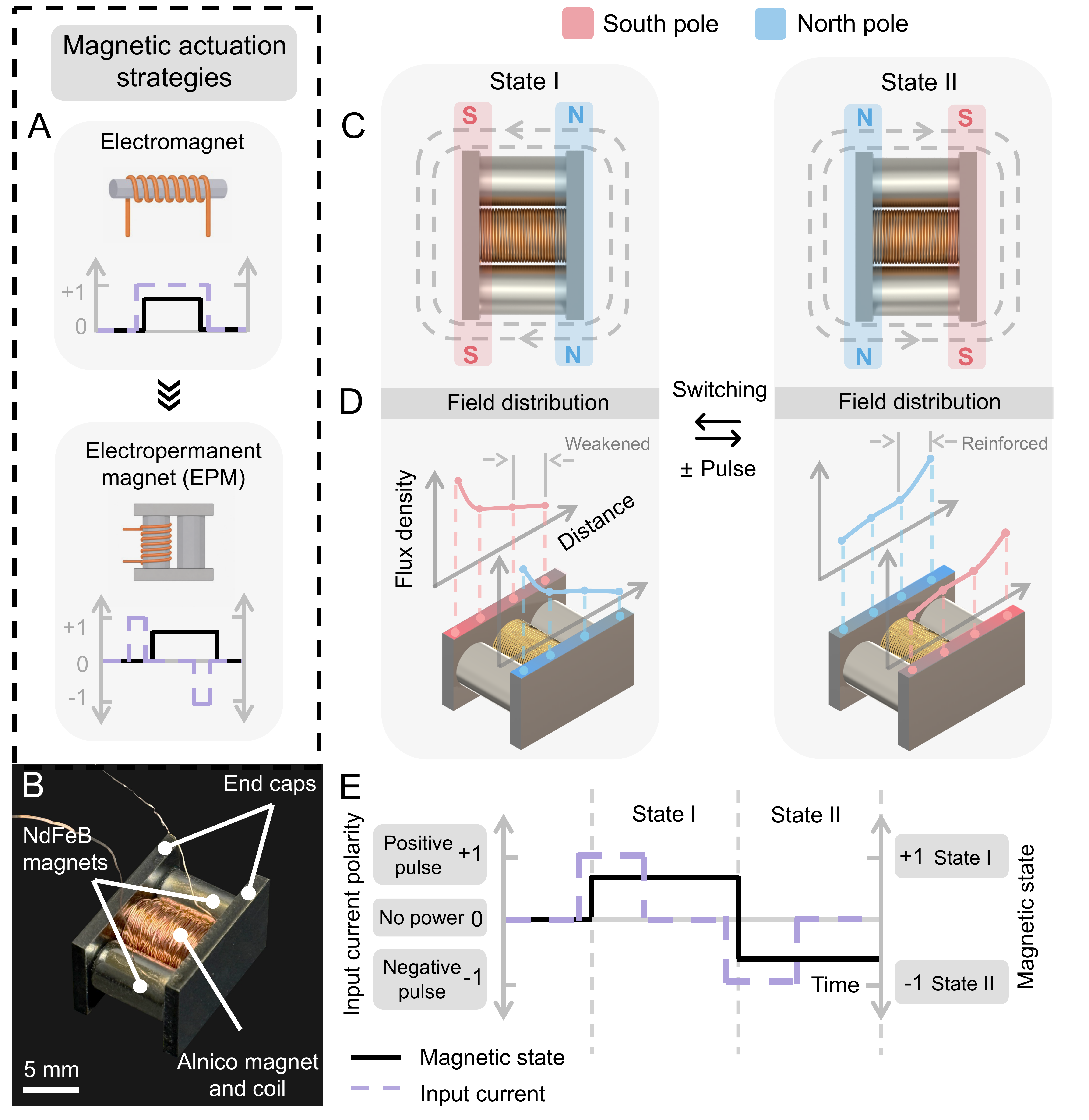}
\captionsetup{font=footnotesize}
\caption{\textbf{Architecture, field distribution, and operating principle of the switchable-polarity electropermanent magnet (S-EPM).}
(A) Comparison of conventional magnetic actuation strategies. An electromagnet requires continuous current, whereas an electropermanent magnet (EPM) is switched by transient current pulses and retains its magnetic state without continuous power input.
(B) Photograph of the fabricated S-EPM, comprising two NdFeB magnets, a coil-wrapped Alnico magnet, and ferromagnetic end caps.
(C) Two bistable states of the S-EPM with reversed external polarity and magnetic flux paths. Red and blue indicate the south and north poles, respectively.
(D) Spatial magnetic-field distributions in the two states. The interaction between the Alnico and NdFeB magnets produces reinforced and weakened field regions, which interchange after polarity switching.
(E) Nonvolatile switching between State I and State II using positive and negative current pulses. After each transient pulse, the selected magnetic state is retained under zero-current conditions.}
\label{fig:S-EPM function}
\end{figure}
To address these limitations, we introduce the Switchable-Polarity Electropermanent Magnet (S-EPM), a bistable magnetic architecture that enables spatial redistribution of the external magnetic field through transient electrical excitation. Unlike standard EPMs that typically switch between oppositely magnetized and demagnetized states, the S-EPM combines a reversibly magnetized low-coercivity Alnico core with adjacent high-coercivity NdFeB magnets to reconfigure internal magnetic flux pathways. Reversal of the Alnico magnetization switches the external magnetic polarity while simultaneously exchanging the reinforced and weakened magnetic-field regions along the ferromagnetic end caps. The resulting magnetic configuration remains stable without continuous power input.

Building upon this S-EPM technology, we developed a programmable valve architecture that routes fluid and switches multiple channels efficiently through five key features: (i) a low-cost, modular architecture using an FDM 3D-printed housing that integrates the S-EPM with an array of permanent magnets; (ii) active bidirectional control enabled by polarity-dependent attraction and repulsion, together with field-matched tube placement that exploits the redistributed magnetic-field strength for enhanced blocking performance; (iii) enhanced sample protection through full fluid isolation and low thermal input, where non-contact, pulse-driven operation minimizes heating and prevents reagent contamination; (iv) high energy efficiency for scale, consuming only 0.6 J per switching event and requiring zero holding power; and (v) broad versatility, supporting pneumatic and hydraulic media while enabling complex combinatorial logic for scalable, autonomous fluidic routing. These characteristics are compared with representative electronically controlled valves in Table 1.

\section{Results}
\subsection*{The Switchable-polarity Electropermanent Magnet (S-EPM)}
Conventional electromagnets require continuous current to sustain a magnetic state, whereas electropermanent magnets (EPMs) use transient pulses to set or cancel a remanent magnetic state without continuous power input (Fig.~1A). Based on this principle, we developed a switchable-polarity electropermanent magnet (S-EPM) that extends conventional EPM operation from on-off activation to control of both the external magnetic polarity and the magnetic-field distribution. The selected magnetic state is retained after the switching pulse is removed, making the operation nonvolatile and eliminating the need for continuous power input. In the S-EPM, current pulses reverse the magnetization of a central Alnico magnet. Its interaction with the fixed NdFeB magnets causes the reinforced and weakened field regions to exchange positions without mechanical reconfiguration. The S-EPM comprises a central Alnico magnet (Alnico 500, equivalent to Alnico 5 grade) wound with a copper excitation coil and flanked by two high-coercivity NdFeB magnets (N38SH) (Fig.~1B). Ferromagnetic end caps enclose the assembly, completing the magnetic flux path and concentrating the external field at the pole interfaces. Detailed fabrication and assembly procedures are provided in Fig.~S2A.

The operating principle of the S-EPM relies on two NdFeB magnets with fixed and opposite axial magnetizations (Fig.~1C and Fig.~S2B). A transient current pulse applied to the coil reverses the magnetization of the Alnico core and reconfigures the magnetic flux distribution. The interaction between the Alnico and NdFeB magnets produces reinforced and weakened external magnetic-field regions on opposite sides of the S-EPM (Fig.~1D). Applying a pulse of opposite polarity reverses the Alnico magnetization and reconfigures the internal flux pathways, causing the reinforced and weakened regions to exchange positions. To experimentally characterize this spatial redistribution, the axial magnetic flux density was measured along the ferromagnetic end-cap edge in both remanent states (Fig.~S3A,B). As a control, the same measurements were repeated using an Alnico-only configuration in which the adjacent NdFeB magnets were removed (Fig.~S3C,D). The S-EPM exhibited a pronounced position-dependent field distribution that reversed after switching, whereas the Alnico-only configuration showed a largely symmetric spatial profile. This comparison confirms that the switchable spatial redistribution of magnetic flux arises from the interaction between the Alnico core and the adjacent NdFeB magnets.
\begin{table}[htbp]
\centering
\scriptsize
\setlength{\tabcolsep}{2.5pt}
\renewcommand{\arraystretch}{1.15}

\begin{tabularx}{\textwidth}
{|>{\raggedright\arraybackslash}X|
c|c|c|c|c|
>{\raggedright\arraybackslash}X|}
\hline

Valve
& \makecell{Mass\\(g)}
& \makecell{Largest\\dimension\\(mm)}
& \makecell{Max. reported\\pressure\\(kPa)}
& \makecell{Driving\\voltage\\(V)}
& \makecell{Working\\medium}
& \makecell{Electrical\\requirement} \\

\hline

Xu et al.\cite{doi:10.1073/pnas.2103198118}
& 5.6
& 40
& 51
& 1200--1800
& Liquid
& 1.1 W (continuous current) \\

Huang et al.\cite{HUANG2022113905}
& 0.88
& 10
& 84
& $\leq 5000$
& Liquid
& N/R (continuous current) \\

Wang et al.\cite{https://doi.org/10.1002/aisy.202200238}
& 30
& 30
& 20
& 12
& Gas
& N/R (continuous current) \\

Leps et al.\cite{Leps_2020}
& 0.476
& 6.5
& 180
& $\sim 12$
& Liquid
& 0.010 J per switch \\

Moran et al.\cite{Moran2024}
& 2.9
& 30
& 146
& 20
& Gas
& 0.047 J per switch \\

Levitronix LPVS-i100 (commercial)\cite{LevitronixLPVSi}
& 390
& 128.9
& $>100$
& 24
& Liquid
& $\sim$1.2 J per switch \\

This work
& 35.5
& 24
& 320
& 48
& Liquid and Gas
& $\sim 0.6$ J per switch \\

\hline
\end{tabularx}

\caption{\textbf{Comparison of electronically controlled valves for fluid-routing and channel-switching applications.} Pressure values correspond to the maximum reported blocking, closing, or regulation pressure under the respective operating and testing conditions. N/R indicates that the electrical requirement was not reported. Values in watts denote continuous power consumption during actuation, whereas values in joules denote the energy required for a single state transition.}

\label{tab:valve-comparison}
\end{table}
After the pulse is removed, the remanence of the Alnico preserves the selected magnetic state without continuous power input (Fig.~1E). The equilibrium between the internal fluid pressure acting on the sealing interface and the magnetic force generated in the magnetized state defines the maximum blocked pressure. We derived the magnetic force following Knaian's analytical framework\cite{Knaian2010}, with the corresponding geometric parameters and lumped magnetic circuit representation defined in Fig.~S6. This framework models the magnetic circuit through Ampere's law and the continuity of magnetic flux:

\begin{equation}
\begin{array}{l}
\frac{\pi d^{2}}{12}\, N_{\mathrm{rods}}
\Big(
  B_{\mathrm{Alnico}}(H_m(t),t)
  + \Delta B_r(t)
  + 2\mu_0 H_m(t)
\Big)
=
\Big(
  \frac{\mu_0 a b}{2g}
  + \mathcalLM{P}_{\mathrm{leak}}
\Big)
\Big(
  N I(t)
  - H_m(t) l
\Big)
\end{array}
\end{equation}

where \( d \) is the diameter of each permanent magnet, and \( N_{\mathrm{rods}} \) is the number of magnetic rods in the S-EPM assembly.
\( B_{\mathrm{Alnico}} \) is the magnetic flux density of the Alnico magnet, which is a nonlinear function of the magnetic field \( H_m(t) \)
and time \( t \) due to hysteresis.
\( B_r \) denotes the remanent flux density of the NdFeB magnets.
The term \( \Delta B_r(t) \) represents the state-dependent effective remanent contribution arising from the asymmetric flux reinforcement between the Alnico and one of the NdFeB magnets, and can be expressed as
\(
\Delta B_r(t) = 2 s_A(t) B_r,
\)
where \( s_A(t) \in \{+1,-1\} \) denotes the magnetization state of the Alnico core.
\( \mu_0 \) is the permeability of free space.
\( a \) and \( b \) represent the width and thickness of the magnetic pole pieces, respectively, while \( g \) is the effective air-gap length in the magnetic circuit.
\( \mathcalLM{P}_{\mathrm{leak}} \) denotes the leakage permeance accounting for flux escaping the principal magnetic path. \( N \) is the number of turns in the excitation coil, \( I(t) \) is the time-dependent current passing through the coil,
and \( l \) is the effective magnetic path length of the magnet stack. To predict the magnetic force generated by the S-EPM during its magnetized state, we express the magnetic pressure in the air gap as:
\begin{equation}
p_m = \frac{B_g^{2}}{2\mu_0}   , \quad
F = 2 p_m a b
= \mu_0 a b \left( \frac{NI(t) - H_m(t) l}{2g} \right)^{2}
\end{equation}
where \( p_m \) denotes the magnetic pressure corresponding to the energy density stored in the air gap. \( F \) represents the magnetic force generated by the S-EPM during its magnetized state. The magnetic energy density method calculates \( F \) by applying the magnetic pressure over the effective pole area (\( a b \)) to yield the total attractive force across the gap. \( B_g \) is the magnetic flux density within the air gap derived from the magnetic-circuit relation. As shown in Equation (2), the magnetic force depends on the geometric dimensions of the S-EPM. The air gap \( g \) plays a dominant role in determining the force magnitude. A more detailed theoretical formulation and derivation of the analytical framework are provided in Supplementary Text section S6. We wound the S-EPM coils with 150 turns of 35 AWG wire to balance energy efficiency with the required magnetic force. This wire gauge and coil geometry lower the electrical resistance, ensuring the necessary transient current is available for fast, reliable switching. We apply a 48 V pulse lasting 1 ms to drive the coil (Fig.~S5). This elevated power rapidly and uniformly reorients the magnetic domains, minimizing the hysteresis delay between the applied field and the resulting magnetization. The optimized design and validated drive method consume approximately 0.6 J per switching event.
\begin{figure}
  \vspace{-1cm}
  \centering
  \includegraphics[width=1\textwidth]{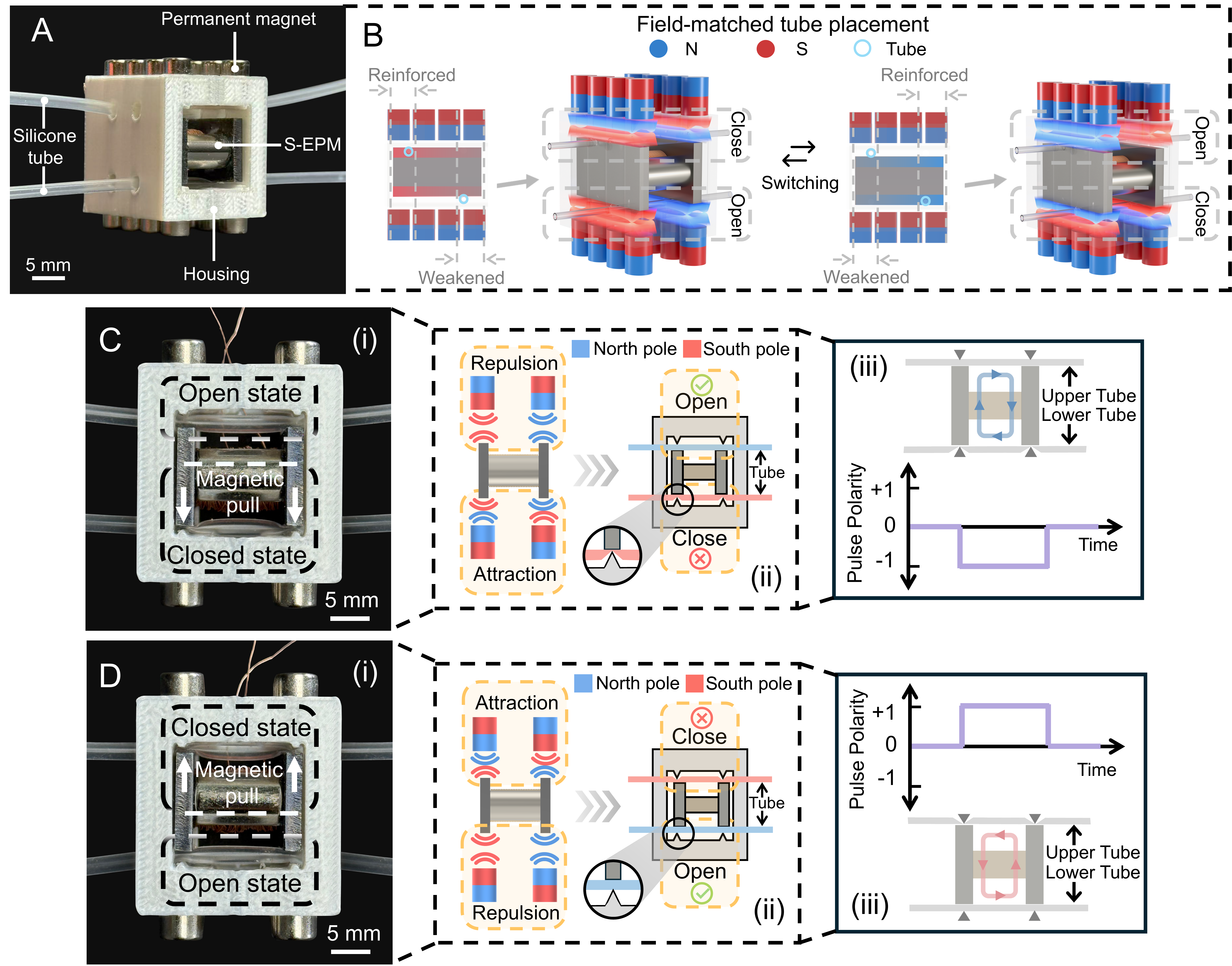}
  \captionsetup{font=footnotesize}
\caption{\textbf{Design and operating principle of the S-EPM valve.}
(A) Photograph of the assembled valve, showing the silicone tubing, 3D-printed housing, central S-EPM actuator, and sixteen permanent magnets arranged in four vertically aligned arrays (four magnets per group).
(B) Field-matched tube placement in the two S-EPM states. The reinforced and weakened magnetic-field regions lie on opposite sides of the actuator. Switching the S-EPM polarity interchanges these regions, so that the tube on the reinforced side is compressed and closed, while the tube on the weakened side remains open. Red and blue indicate the south and north poles, respectively, and the cyan circles indicate the tube positions.
(C) Downward actuation state. (i) Photograph of the valve with the lower tube compressed (closed) and the upper tube open. (ii) Schematic of the magnetic configuration with a magnified view of the lower tube deformation, highlighting compression-induced occlusion. (iii) Corresponding excitation pulse (pulse polarity = $-1$) and clockwise magnetization of the Alnico core.
(D) Upward actuation state. (i) Photograph of the complementary configuration with the upper tube closed and the lower tube open. (ii) Schematic of the alternate magnetic configuration with a magnified view of the uncompressed lower tube. (iii) Corresponding excitation pulse (pulse polarity = $+1$) and counterclockwise magnetization of the Alnico core.}
  \label{fig:data}
\end{figure}

\subsection*{S-EPM valve design}

The S-EPM valve integrates a bistable magnetic actuator within a compact pinch-valve architecture to switch deterministically between two fluidic channels (Fig.~2A and Movie S1). The assembly comprises four primary components: (i) the S-EPM actuator, (ii) a rigid PLA housing fabricated via FDM 3D printing, (iii) two parallel silicone tubes (outer diameter 2 mm, inner diameter 1.5 mm), and (iv) sixteen static N40 permanent magnets (5 mm diameter, 7 mm length) arranged in four vertically aligned groups around the actuator. Magnets positioned along opposing vertical axes present opposite poles toward the S-EPM, establishing a vertical magnetic force gradient when the actuator polarity reverses. The housing incorporates two raised internal bars on both the upper and lower inner surfaces, aligned with the S-EPM end caps. These protruding features concentrate the contact area between the S-EPM and the silicone tubing. This design increases the local compressive stress and ensures reliable tube occlusion under the available magnetic force.

A transient voltage pulse (1 ms) applied to the coil reverses the magnetization of the Alnico core, reconfiguring both the external magnetic polarity and the spatial magnetic-field distribution of the S-EPM (Fig.~2B). This switching causes the reinforced and weakened magnetic-field regions to exchange positions. At the same time, reversal of the external polarity changes the attraction and repulsion between the S-EPM and the surrounding permanent-magnet arrays, reversing the direction of the net magnetic force and switching the actuator between the two tube positions. The spatial variation in field strength determines the magnitude of the magnetic interaction at each side of the actuator and therefore influences the resulting blocking performance.

In the configuration that Fig.~2C(i) shows, the lower tube is compressed (closed) while the upper tube remains open. As Fig.~2C(ii) illustrates, the lower magnet array attracts the S-EPM and the upper array repels it, producing a net downward magnetic force. This force drives the actuator downward, compressing the lower silicone tube and blocking flow. Simultaneously, it releases the upper channel. A negative excitation pulse (pulse polarity = $-1$) establishes this magnetic configuration by setting the corresponding remanent magnetization state of the Alnico core (Fig.~2C(iii)).

Reversing the excitation pulse produces the complementary configuration that Fig.~2D shows. In Fig.~2D(i), the upper tube is compressed while the lower tube remains open. As Fig.~2D(ii) depicts, the system attracts the S-EPM upward and repels it downward, generating a net upward force that displaces the actuator and seals the upper channel. A positive excitation pulse (pulse polarity = $+1$) establishes this complementary remanent state of the Alnico core (Fig.~2D(iii)).

Because of the remanent magnetization of the Alnico core, each actuation state remains mechanically stable after removal of the excitation pulse, enabling bistable operation without continuous power input. We evaluate valve performance by measuring the maximum blocked pressure, defined as the critical internal pressure the valve can withstand without leakage.

\begin{figure}
  \centering
  \includegraphics[width=1\textwidth]{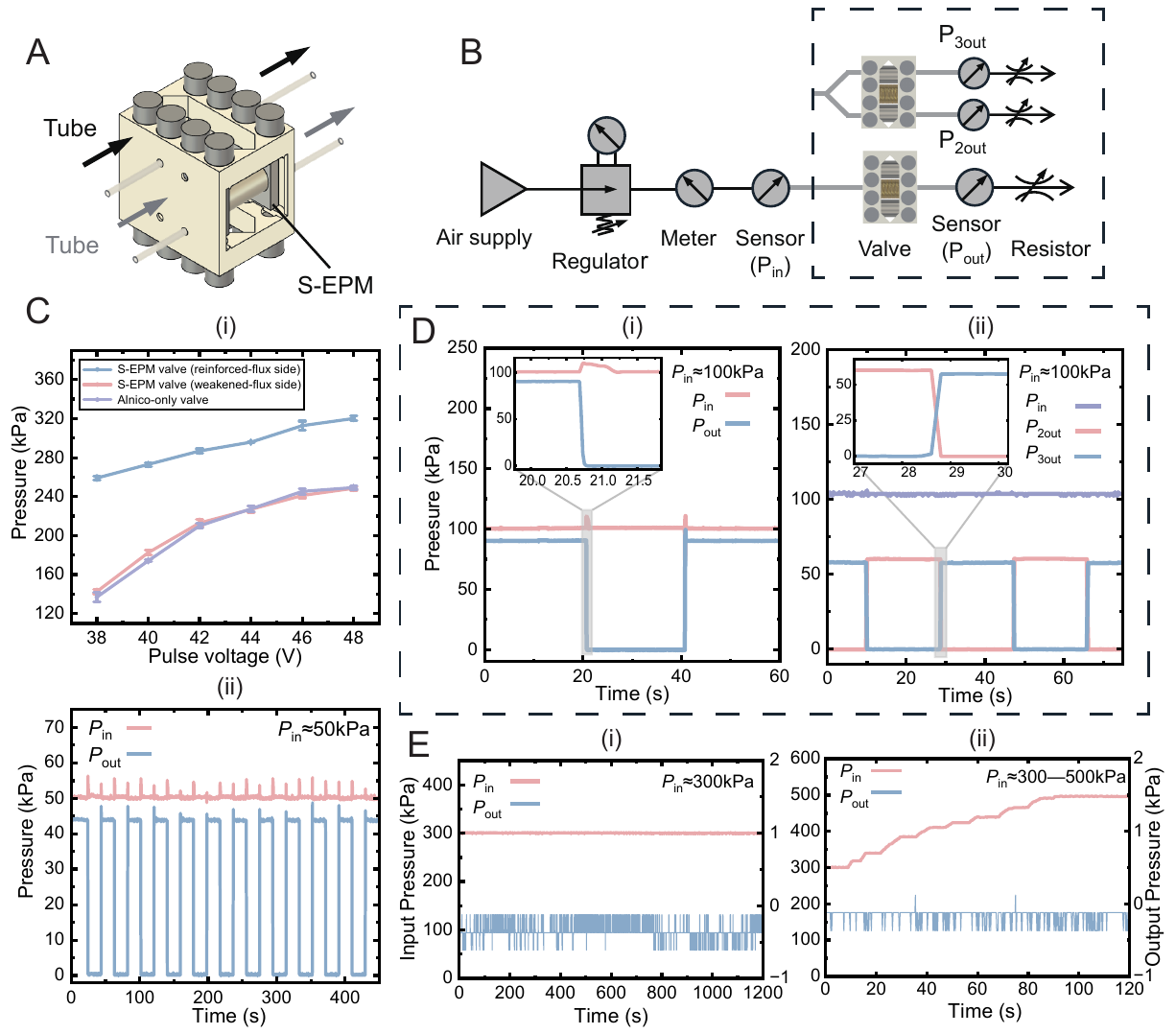}
  \captionsetup{font=footnotesize}
  \caption{\textbf{Pressure blocking characterization of the S-EPM valve.}
(A) Photograph of the S-EPM valve integrated with compressible tubing.
(B) Experimental setup for pressure-blocking measurements, including regulated air supply, flow meter, upstream ($P_{in}$) and downstream ($P_{out}$) pressure sensors, and venting resistor.
(C) Dynamic blocking performance: (i) Dynamic blocking pressure at different pulse voltages for the S-EPM valve with the tube positioned on the reinforced and weakened sides, compared with the Alnico-only valve. Data are presented as mean ± SD (n=3 for each S-EPM condition and n=5 for the Alnico-only valve); (ii) cyclic switching at $P_{in}\approx50$ kPa.
(D) Transient switching dynamics at $P_{in}\approx100$ kPa:
(i) time-resolved single-channel switching transient;
(ii) dual-outlet alternating operation showing complementary pressurization profiles.
(E) Static sealing tests: (i) long-duration blocking at $P_{in}\approx300$ kPa for 20 min, with stable upstream pressure and the secondary y-axis resolving the low-pressure $P_{out}$ signal; (ii) stepwise upstream pressure ramp from 300 to 500 kPa with corresponding $P_{in}$ and $P_{out}$ traces.}
  \label{fig:pressure_data}
\end{figure}

\subsection*{Characterizing the pressure blocking capability of the S-EPM valve}

To integrate the S-EPM valve effectively into configurable fluidic systems, we characterized the device across five performance metrics: dynamic blocking capacity, cyclic switching reliability, switching dynamics, temporal sealing stability, and static pressure limits. We adapted the experimental apparatus for evaluating blocking performance from Moran's methodology \cite{Moran2024} (Fig.~3A). We directed compressed air from a regulated supply through a flow meter and an upstream pressure sensor ($P_{in}$) to the valve inlet. We connected the valve outlet to a downstream pressure sensor ($P_{out}$) and a high-resistance venting element, allowing controlled depressurization to the atmosphere (Fig.~3B). This configuration enabled precise quantification of the differential pressure maintained across the valve in the blocked state.

We characterized the switching performance of the S-EPM valve by analyzing pressure transients as the system toggled the measured channel between open and closed states. We aimed to verify the valve capacity to isolate the inlet from the outlet. This isolation maintains pneumatic energy in downstream actuators and prevents unintended dissipation. Upon actuation, the valve exhibited distinct flow behaviors. During the closed-to-open transition, the downstream region rapidly equilibrated with the supply pressure, ensuring effective flow transmission. Conversely, the open-to-closed transition triggered a near-instantaneous venting of the outlet to ambient levels while the inlet pressure remained stable. These switching dynamics allow the valve to block flow selectively and maintain high differential pressures.

To evaluate how the spatial magnetic-field distribution affects valve performance, we compared the S-EPM valve with an Alnico-only control valve. In the control configuration, the adjacent NdFeB magnets were removed while the switchable Alnico magnet, excitation coil, ferromagnetic end caps, and valve architecture were retained (Fig.~S9A,B). Dynamic blocking pressure was measured over a pulse-voltage range of 38 to 48 V (Fig.~3C(i)). For all configurations, the blocking pressure increased with pulse voltage. The S-EPM valve achieved substantially higher blocking pressures when the tube was positioned on the reinforced-flux side, reaching approximately 320 kPa at 48 V. In contrast, the weakened-flux side of the S-EPM and the Alnico-only valve exhibited similar, lower blocking pressures over the tested voltage range. Measurements extending to lower pulse voltages further showed that reliable switching of the Alnico-only valve was obtained from approximately 38 V (Fig.~S9C). These results show that the enhanced blocking capability is associated with the reinforced magnetic-field region generated by the interaction between the Alnico core and the adjacent NdFeB magnets.

To further evaluate dynamic blocking performance, additional tests were conducted near the maximum dynamic blocking pressure of approximately $320$ kPa, at inlet pressures of approximately $310$, $315$, and $320$ kPa (Fig.~S7A), together with tests at intermediate pressures of approximately $50$, $125$, and $200$ kPa (Fig.~S7B). In all cases, valve closure rapidly reduced the outlet pressure to ambient levels while the inlet pressure remained stable, demonstrating consistent blocking performance across the tested pressure range. This consistent, leak-free response at pressures up to $320$ kPa demonstrates the robust sealing integrity for high-pressure applications. We assessed operational reliability through cyclic switching tests at a nominal pressure of $\approx 50$ kPa (Fig.~3C(ii)). These experiments yielded highly reproducible pressure profiles. The downstream region promptly equilibrated with the inlet during opening and vented rapidly during closing. We observed no signal drift or amplitude degradation over repeated cycles, confirming consistent flow-blocking and transmission performance under frequent actuation.

High-resolution analysis at an inlet pressure of approximately 100 kPa quantified the dynamic response speed. The transition from the open to the closed state depressurized the outlet within 0.115 s (Fig.~3D(i)). This rapid closure induced a momentary pressure spike at the inlet due to the sudden arrest of flow. The spike quickly dissipated as the system returned to steady state, validating the swift closure mechanism without compromising upstream stability. To characterize the switching latency between alternate flow paths, we employed a dual-outlet experimental setup (Fig.~3D(ii)). The recorded pressure profiles displayed a synchronized alternating pattern; the pressurization of one outlet channel coincided with the venting of the other. The crossover transition occurred within a $< 0.5$ s window. This verifies the intrinsic bistable operation of the S-EPM, where only one fluidic path remains active at any given time.

Static tests conducted over extended durations and at elevated pressures further investigated leakage behavior in the closed state. As Fig.~3E(i) shows, we subjected the valve to a constant supply pressure of $\approx 300$ kPa for $20$ minutes. The outlet pressure remained at ambient levels with negligible fluctuations. This demonstrates that leakage does not accumulate over time and the blocking function remains stable throughout extended operation. Furthermore, as Fig.~3E(ii) illustrates, we incrementally increased the upstream pressure from $300$ to $500$ kPa. Despite this substantial increase in differential pressure, the outlet trace showed no discernible rise. The sealing integrity of the valve remains insensitive to inlet pressure variations within this regime. These results confirm that the valve maintains a leakage-free state without performance deterioration, even when pushed to the $500$ kPa upper limit of the testing apparatus.

For comparison, the static blocking capability of the Alnico-only control valve was evaluated over a pulse-voltage range of $38$ to $48$ V (Fig.~S9D). Using an outlet-pressure threshold of $3$ kPa to identify the onset of leakage, the static blocking pressure increased with pulse voltage and reached approximately $370$ kPa at $48$ V. A representative measurement near this limit is shown in Fig.~S9E, where an input pressure of approximately $375$ kPa resulted in an outlet pressure of approximately $3.54$ kPa, exceeding the defined leakage threshold. Compared with the Alnico-only control, the S-EPM valve therefore provides a higher static sealing capability, consistent with the enhanced dynamic blocking performance observed in Fig.~3C(i).

Despite the higher pressure sustained under static sealing conditions, the dynamic blocking pressure reached a practical limit of approximately $320$ kPa. This difference arises because dynamic closure requires the actuator to displace and fully compress the pressurized silicone tube during switching, whereas static sealing is evaluated after the tube has already been occluded. Near the dynamic limit, the magnetic interaction generated in the selected remanent state occasionally failed to complete tube compression in a single actuation, and additional pulses were required to establish a stable seal. This behavior indicates that the dynamic threshold is governed by the available actuation force during closure rather than by the pressure-bearing capability of the already sealed valve. Further optimization of the magnetic interaction, actuator positioning, and switching conditions could therefore improve reliable dynamic operation at higher pressures.

\subsection*{Multi-Modal Fluidic Control Architectures enabled by S-EPM Valves}
Laboratory automation requires programmable fluid handling. Experimental workflows routinely selectively deliver, mix, and route liquids among multiple reservoirs, reaction sites, and output channels \cite{D4LC00423J,Werner2021ProgrammableDroplet}. Current approaches rely on numerous individual valves and extensive external pumping infrastructures. These resulting complex control architectures face significant scalability challenges as system size grows \cite{Brower2017}.

At the system level, we can abstract many laboratory operations as address-based fluid selection and routing tasks, analogous to the operation of decoders and multiplexers in digital logic. Translating this abstraction into the fluidic domain remains challenging. Standard valve-based control architectures introduce substantial integration, actuation, and energy overhead as the number of controlled elements increases \cite{D4LC00423J,Wang2021CPCVReview}.

To address these limitations, we realized decoder- and multiplexer-like logic within fluidic networks of S-EPMs. This integrated S-EPM combination enables address-based routing and distribution without continuous actuation. It leverages the inherent bistability of S-EPM technology.
\begin{figure}
  \centering
  \includegraphics[width=1\textwidth]{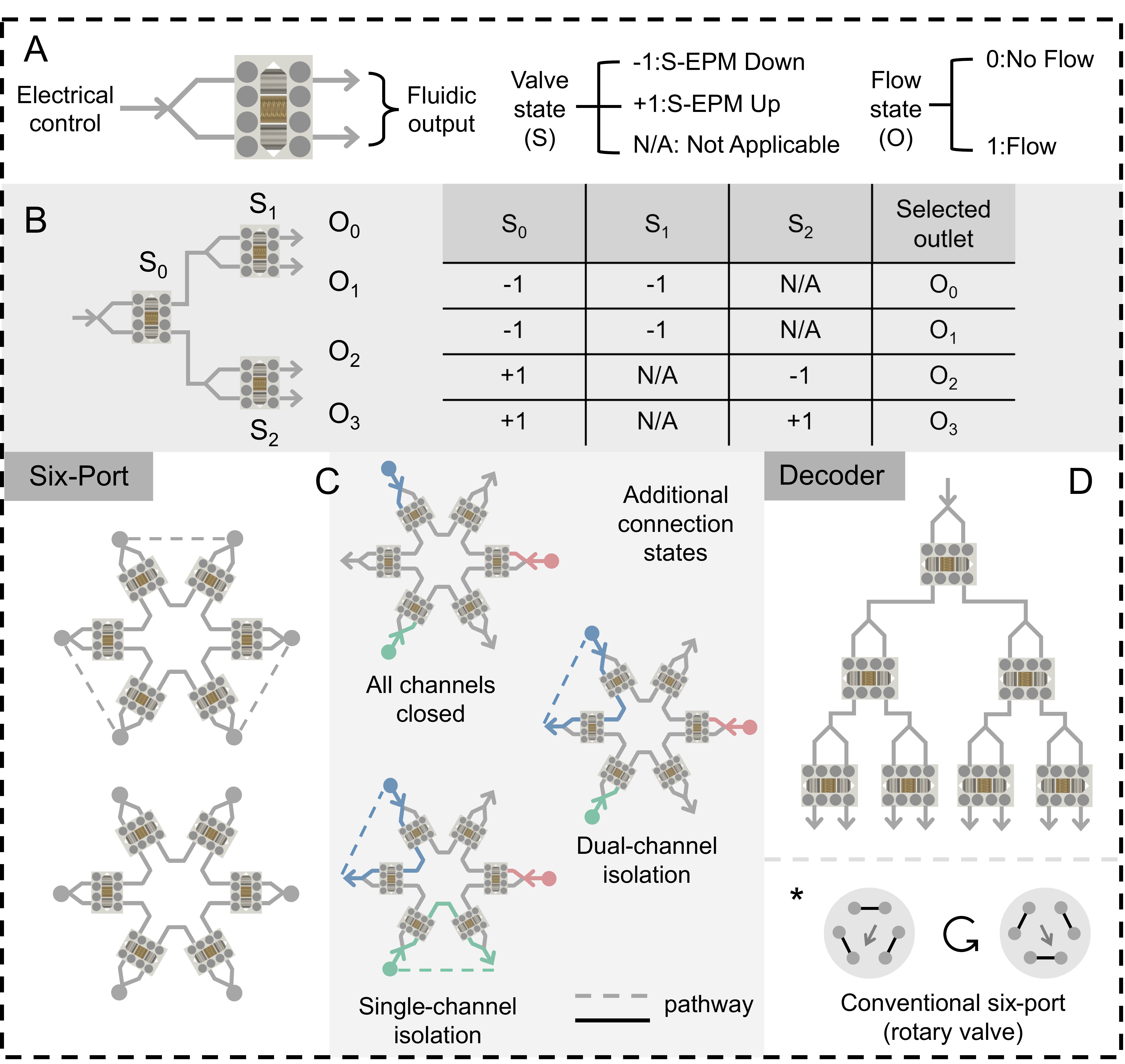}
  \captionsetup{font=footnotesize}
  \caption{\textbf{Architectural configurations of S-EPM-based fluidic routing.}
(A) State-controlled routing unit. Electrical control sets the bistable state ($S=-1$ or $S=+1$) of an S-EPM valve, which determines the corresponding fluidic output state ($O$), defined as $O=0$ for no flow and $O=1$ for flow.
(B) Hierarchical tree-like routing module. Three S-EPM valves with states $S_0$--$S_2$ direct a single input to one of four outlets ($O_0$--$O_3$). The table shows the mapping between valve-state combinations and the selected outlet, where N/A indicates that the corresponding valve state does not affect the selected route.
(C) Six-port configuration with independently programmable connections between adjacent ports. In addition to the two alternating six-port connection patterns, the S-EPM architecture supports additional states including all channels closed, single-channel isolation, and dual-channel isolation. A conventional mechanically actuated six-port rotary valve is shown for comparison, with its connections restricted to fixed routing patterns.
(D) Decoder configuration in which combinations of S-EPM valve states select a specific fluidic outlet from a single input.
}

  \label{fig:arcihtectures}
\end{figure}
% Architecture figure
We integrated S-EPM valves into diverse architectural configurations to achieve low-power, combinatorial fluidic routing (Fig.~4 and Fig.~S4), including representative dual-valve and triple-valve implementations (Movie S2). At the fundamental level (Fig.~4A), each S-EPM valve functions as a nonvolatile state-controlled fluidic element. A transient electrical input sets the bistable valve state, denoted by $S=-1$ or $S=+1$, corresponding to the two stable S-EPM positions. The resulting fluidic state is described by $O=0$ for no flow and $O=1$ for flow. Once programmed, the valve maintains its state without continuous electrical power.

Building on this state representation, we implemented a hierarchical tree-like routing topology (Fig.~4B). In the four-output configuration, three S-EPM valves with states $S_0$--$S_2$ route a single fluidic input to one of four outputs ($O_0$--$O_3$) through successive branching decisions. The state of $S_0$ first selects one of the two branches, after which either $S_1$ or $S_2$ determines the final output. The truth table summarizes these routing combinations, with N/A indicating that the valve state in the unselected branch does not affect the resulting output.

% ===== Figure 5 image =====
\begin{center}
  \vspace*{1cm}
  \includegraphics[width=1\textwidth]{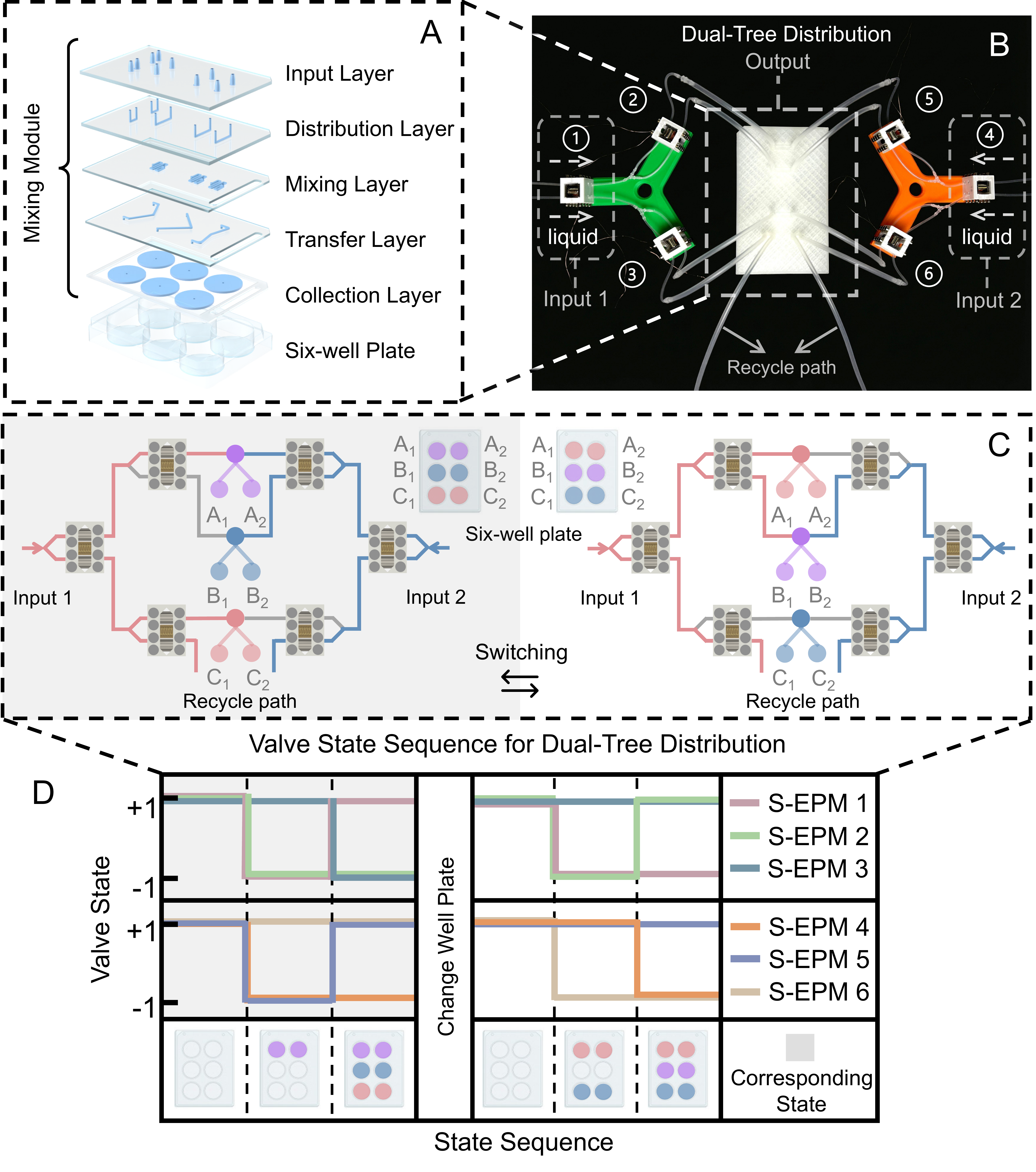}
\end{center}

% If there is not enough space left for the caption,
% move the caption to the next page as a whole.
\Needspace{14\baselineskip}

\captionsetup{type=figure,font=footnotesize}
\captionof{figure}{
\textbf{Nonvolatile dual-tree distribution and mixing via cooperative S-EPM valve modules.}
(A) Layered architecture of the FDM-fabricated mixing module, comprising input, distribution, mixing, transfer, and collection layers for vertically integrated flow management, with outputs directed into a six-well plate.
(B) Physical implementation of the dual-tree mixer, featuring two symmetric three-valve S-EPM modules coupled to a central mixing unit with a shared downstream output.
(C) Reconfigurable mixing behavior in the dual-tree architecture demonstrated using red and blue dyes as representative inputs. Upon interaction, the two streams produce a purple output, enabling direct visualization of mixing. By switching the valve states, the system transitions between two configurations, resulting in a shift of the mixing region across the six-well plate. The labeled nodes (A1--C2) correspond directly to individual wells, highlighting programmable control over both routing and the spatial location of mixing, with excess flow directed through defined recycle paths.
(D) Representative valve state sequence governing dual-tree operation. Values +1 and $-1$ denote the two stable states of each S-EPM valve. Each column corresponds to a distinct system configuration, with the well-plate schematics indicating the resulting distribution pattern. The grey panels in (D) correspond to the grey-highlighted distribution states shown in (C).
}
\label{fig:mixer}

\vspace{0.8em}

\subsubsection*{A Nonvolatile Dual-Tree distribution via cooperative S-EPM valve modules}

Modular S-EPM valve groups coordinate to achieve Dual-Tree distribution (Fig.~5C). The physical implementation features two identical valve modules positioned symmetrically around a central mixing unit (Fig.~5B). Each module comprises three S-EPM valves with a single input and four controllable output branches. While these two modules are independently addressable, they fluidically couple to the same downstream mixing module.

\hspace*{2em}To enable versatile liquid handling, the FDM 3D-printed central mixing module uses a stacked configuration of internal flow channels (Fig.~5A). This architecture allows the system to combine reagents from the upstream valves or route them separately before they exit through the bottom. By positioning a six-well plate beneath the module and reprogramming the S-EPM valve states, the system dynamically determines the distribution pattern. This architecture provides a choice between pristine reagent delivery or on-chip mixing.

\hspace*{2em}The functional versatility of this distribution logic is demonstrated using red and blue dyes as representative inputs (Fig.~5C). Combining the dyes produces a purple output, visually confirming successful mixing. The system facilitates programmable one-to-many distribution. In one specific configuration, certain wells receive mixed reagents while others receive individual dyes. Upon transitioning to an alternate state, the system reconfigures these roles. This outcome demonstrates that the physical architecture supports multiple, distinct distribution topologies.

\hspace*{2em}A representative valve state sequence demonstrates the logic-driven nature of these transitions (Fig.~5D and Movie S3). The values +1 and -1 denote the two stable states of each S-EPM valve. We organize the sequence into two rows (Valves 1--3 and 4--6), with each column representing a discrete system configuration. The well-plate schematics at the base of each column confirm the resulting reagent distribution for that state. This representation shows that each distribution configuration is defined by the programmed valve states, highlighting the architectural flexibility of the S-EPM-based design.

\subsubsection*{Six-port routing via modular S-EPM valve arrays}

While the cooperative modules excel at one-to-many distribution, we also reconfigured the modular S-EPM architecture into a circular array to replicate and enhance current routing paradigms. The six-port rotary valve is a staple of fluid routing in chemistry and life-science laboratories. It enables deterministic switching between predefined flow paths. Although widely employed in automated systems, such as the platform demonstrated by Steiner \cite{steiner_organic_2019}, current rotary valves rely on bulky electromechanical actuation and suffer from mechanically coupled flow paths, moving parts, high cost, and limited repairability.

\hspace*{2em}Implementing a six-port routing architecture through a circular array of modular S-EPM valves (Fig.~4C) reproduces the two standard connection patterns of a conventional six-port rotary valve while expanding the accessible routing states. In conventional rotary valves, the port connections are mechanically coupled and therefore switch collectively between predefined configurations. In contrast, each S-EPM valve in the circular array can be programmed independently. This independent addressability enables additional connection states beyond the two conventional patterns, including all channels closed, single-channel isolation, and dual-channel isolation. Individual pathways can therefore be selectively retained or isolated without requiring the remaining ports to follow a mechanically linked switching sequence.

% ===== Figure 6 image =====
\begin{center}
  \vspace*{-2cm}
  \includegraphics[width=\textwidth]{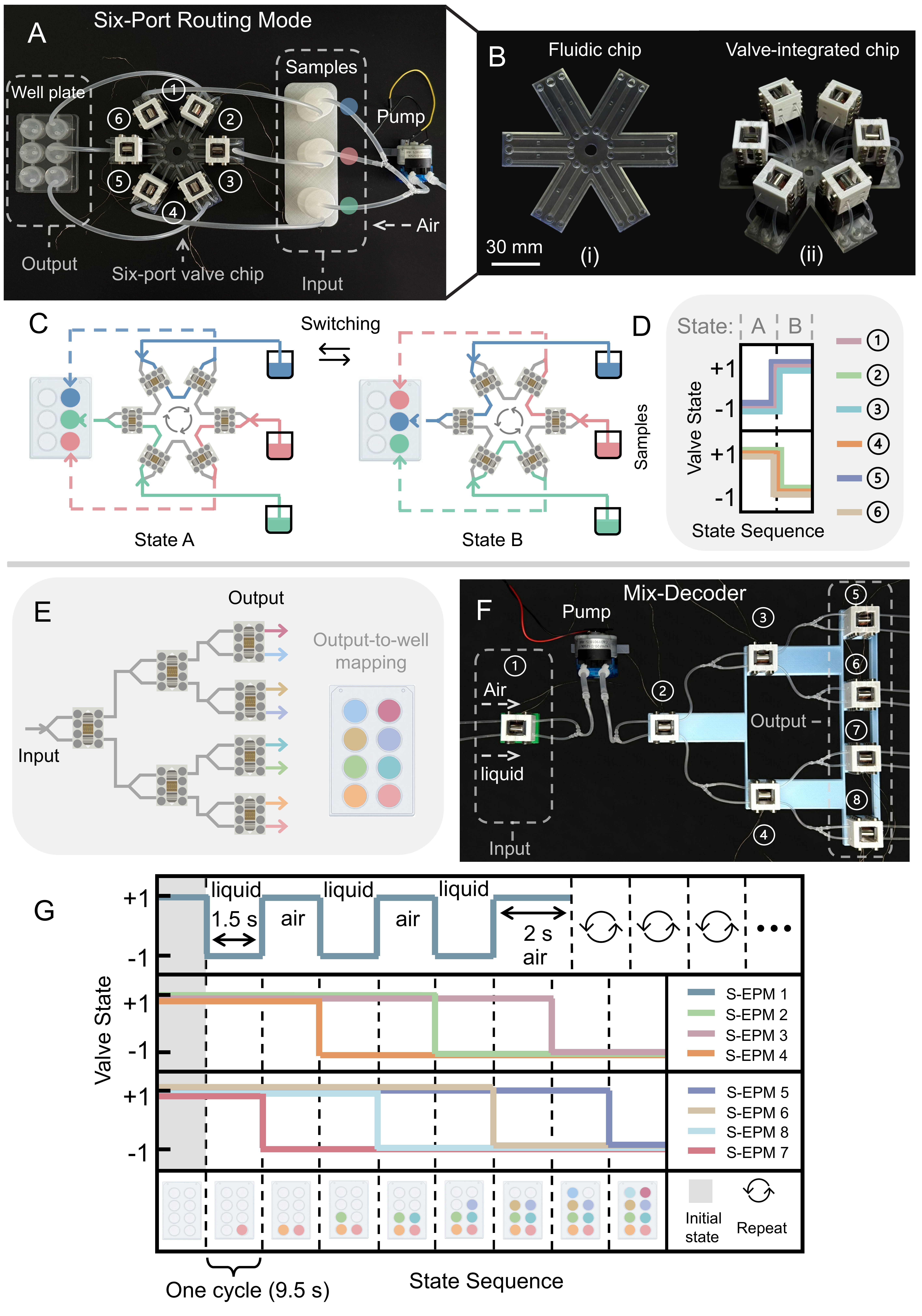}
\end{center}

\newpage

% ===== Figure 6 caption =====
\captionsetup{
  type=figure,
  font=footnotesize
}

\captionof{figure}{
\textbf{Reconfigurable six-port routing and mix-decoder operation using modular S-EPM valves.}
(A) Experimental setup for six-port routing, with three fluid inputs connected to a six-port valve chip and directed to a well plate under pressure supplied by a single pump.
(B) Six-port device comprising (i) the fluidic chip and (ii) the valve-integrated chip with six S-EPM valves.
(C) Two programmed routing states of the six-port valve chip. Switching the valve states redirects the three input fluids between adjacent output pathways.
(D) State sequence of the six S-EPM valves corresponding to States A and B in (C), where $+1$ and $-1$ denote the two stable valve states.
(E) Tree-structured decoder showing the one-to-one mapping between eight fluidic outputs and the corresponding wells.
(F) Experimental mix-decoder setup integrating input selection and downstream routing using eight S-EPM valves.
(G) Time-sequenced operation of the mix-decoder. Alternating liquid and air delivery is combined with programmed S-EPM state changes to sequentially address the output wells. The well-plate schematics show the resulting distribution after each step of a 9.5~s cycle, which can be repeated for continued operation.
}

\label{fig:6port}

\vspace{0.9em}
\hspace*{2em}Building on this architecture, we constructed the six-port routing system shown in Fig.~6A. Three independent sample streams are supplied to alternating ports of the circular module and driven pneumatically by a single external pump, while the routed fluids are collected in a well plate. The underlying six-branch fluidic chip and the fully assembled valve-integrated device are shown in Fig. 6B, with additional details of the fluidic layout and the integration between the S-EPM valves and the chip provided in Fig. S8. Six S-EPM valves are distributed around the central junction, allowing the connectivity between neighboring ports to be reconfigured through their independently programmed bistable states while the physical chip and tubing arrangement remain unchanged.

\hspace*{2em}A representative routing demonstration is presented in Fig.~6C and Movie S4. In State A, the device establishes three parallel pathways, corresponding to ports 1 to 2, 3 to 4, and 5 to 6, delivering the three sample streams to separate outputs. Switching the valve states reconfigures the same device into State B, with connections from ports 1 to 6, 3 to 2, and 5 to 4, thereby redirecting each input toward the alternate neighboring outlet. The corresponding state transition of the six S-EPM valves is summarized in Fig.~6D, where $+1$ and $-1$ denote their two stable magnetic states. This demonstration verifies that the six-port routing topology can be reconfigured through brief electrical programming without mechanical rotation or modification of the fluidic hardware.

\subsubsection*{Decoder operation based on modular S-EPM valves}
We constructed a fluidic decoder by arranging S-EPM valves into a tree-like network. This network selectively routes a single input to one of multiple output branches through discrete valve state combinations (Fig.~4D and Movie S5). In this configuration, seven S-EPM valves address eight independent output paths, routing a single fluidic input to a selected output according to the programmed valve-state combination. Interpreted inversely, the same architecture functions as a multiplexer. It selects one branch among multiple candidates to connect to a shared channel, depending on the programmed valve states. To integrate distribution and compositional control within this network, we developed the eight-valve mix-decoder module (Fig.~6F). This physical realization uses a single external pump to supply various fluid media under continuous driving pressure. The system governs the selection and routing of each medium solely through encoded S-EPM valve states. This enables seamless transitions between simple distribution and complex, compositionally controlled routing without modifying the physical layout. The output-to-well mapping is illustrated in Fig.~6E, where each decoder output is assigned to a designated well in the well plate. This provides a direct visual mapping between routing states and spatially addressed wells.

\hspace*{2em}The valve state sequence (Fig.~6G and Movie S5) demonstrates the dynamic execution of this mix-decoder logic. In this example, we supply two distinct fluid media to the input. S-EPM~1 switches between logical states +1 and -1 to select the delivered medium at each step. This programmed sequence alternates between liquid and gas over multiple 1.5~s intervals, followed by a final 2~s gas phase. Together, they form a complete 9.5~s control cycle.

\hspace*{2em}After each cycle, the system sequentially advances the decoder valve states to redirect the input stream to a subsequent output branch. Each 9.5~s cycle targets a single well of the well plate. Repeating this process addresses all eight wells using identical control logic and timing. This operation demonstrates how integrating temporal encoding at the input with spatial decoding enables the programmable distribution of mixed media. In the routing demonstrations, +1 and -1 denote the two stable S-EPM valve states rather than the instantaneous switching pulse polarity.

\hspace*{2em}These demonstrations establish a unified framework for logic-inspired fluidic control. Discrete valve states, rather than fixed plumbing, govern routing, distribution, and compositional operations. By reprogramming valve states within a fixed fluidic topology, complex routing workflows can be reconfigured without modifying the physical hardware. Importantly, this architecture replicates and extends classical components such as the six-port rotary valve by introducing independent path selection and persistent operation without sustained power input. Consequently, systems enabled by the S-EPM provide a scalable, energy-efficient, and highly reconfigurable pathway toward next-generation programmable laboratory automation. Additional application-oriented demonstrations of stacked S-EPM valves for wearable pneumatic assistance and thermal-fluid routing are provided in Supplementary Text section S1, Fig.~S1, and Movies S6 and S7. These supplementary examples are included to illustrate the broader application versatility of the S-EPM valve, while the main system-level demonstrations focus on programmable fluidic routing, distribution, and decoding architectures presented above.

\section{Discussion}

The S-EPM extends electropermanent magnetic actuation from remanent-state switching to programmable redistribution of the external magnetic field. By coupling a switchable Alnico core with fixed NdFeB magnets, reversal of the Alnico magnetization reconfigures the internal flux pathways and exchanges the reinforced and weakened field regions along the end-cap direction. This builds on the nonvolatile magnetic-state control of conventional electropermanent magnets\cite{Knaian2010}, while introducing spatial control over the field distribution. In comparison, the Alnico-only configuration produces a largely symmetric field profile, confirming that the adjacent NdFeB magnets are responsible for the directional redistribution observed in the complete S-EPM. This redistribution translates directly into improved valve performance, with a maximum dynamic blocking pressure of approximately $320$ kPa and static sealing of at least $500$ kPa, exceeding the pressure capability reported for several dielectric- and hydrogel-based actuation approaches\cite{doi:10.1073/pnas.2103198118,article_ionov}.

\hspace*{2em}The S-EPM valve also retains the practical advantages of nonvolatile magnetic actuation. Unlike conventional solenoid valves, which require sustained electrical power and can generate Joule heating\cite{kim2012lifting}, the S-EPM requires electrical input only during switching and maintains its selected state without continuous current. Independently retained valve states further enable programmable fluidic networks within fixed hardware. The hierarchical distribution, decoder, mix-decoder, and six-port demonstrations show that routing configurations can be reprogrammed without continuous actuation, reducing reliance on external control hardware that becomes increasingly difficult to scale with channel number\cite{Brower2017,B820557B}. In the six-port architecture, independent valve addressability also enables routing states beyond the mechanically coupled configurations of conventional rotary valves.

\hspace*{2em}The present valve has a lower dynamic blocking capability than its static sealing capability because dynamic closure requires the actuator to compress an already pressurized tube. Further optimization of the magnetic circuit, actuator geometry, coil design, and excitation conditions could increase the available switching force and support further miniaturization. Overall, the S-EPM combines persistent magnetic-state control with spatial field redistribution, enabling both enhanced valve performance and programmable routing within fixed fluidic architectures. This architecture provides a basis for scalable, low-power fluidic systems in laboratory automation, soft robotics, and other applications requiring persistent and reconfigurable flow control.

\section{Materials and Methods}
\subsection*{S-EPM Fabrication}

The Switchable-polarity Electropermanent Magnet (S-EPM) comprises a central low-coercivity Alnico 500 magnet surrounded by an excitation coil and flanked by two high-coercivity NdFeB magnets (N38SH) with opposing axial magnetizations. Ferromagnetic iron end caps enclose the assembly to complete the magnetic circuit and concentrate the external field at the pole interfaces.

We wound the excitation coil with 150 turns of 35 AWG copper wire. We achieved polarity switching using a single transient voltage pulse (48 V, 1 ms). This pulse reverses the remanent magnetization of the Alnico core and thereby toggles the external polarity. Each switching event consumed $0.6\,\mathrm{J}$.

The supplementary materials provide detailed fabrication procedures and assembly steps.

\subsection*{Valve Assembly}

The S-EPM valve integrates the bistable actuator within a compact pinch-valve architecture. We fabricated the housing using fused deposition modeling (FDM) 3D printing in PLA. We positioned two parallel silicone tubes (outer diameter 2 mm, inner diameter 1.5 mm) between the actuator end caps and internal raised support features to enable localized magnetic compression.

We arranged sixteen N40 permanent magnets (5 mm diameter $\times$ 7 mm length) in four vertically aligned arrays surrounding the actuator. This generated a polarity-dependent magnetic force gradient. Magnets positioned along opposing vertical axes presented opposite poles toward the S-EPM, producing bidirectional actuation under polarity reversal.

The remanent polarity of the S-EPM ($+1$ and $-1$) defined the valve states, independent of the driving pulse once switching concluded.

\subsection*{Electrical Actuation}

We generated transient excitation pulses using an H-bridge driver controlled by a microcontroller. Each switching event consisted of a single 1 ms voltage pulse (48 V). The pulse polarity determined the final remanent state of the S-EPM.

We applied no continuous holding current; magnetic remanence alone maintained the valve states. For multi-valve routing and decoder demonstrations, we addressed individual S-EPM units independently according to predefined state sequences.

The supplementary materials provide a detailed description of the electrical setup, including the driving circuitry and control architecture.

\subsection*{Pressure Characterization}

We evaluated pressure blocking performance using regulated compressed air. We measured upstream ($P_{\mathrm{in}}$) and downstream ($P_{\mathrm{out}}$) pressures using amplified analog pressure transducers (Nidec P-7100 series).
For low-pressure and compound measurements ($-100$ to $100$~kPa), we used a P-7100-102R-M5 sensor (0.5--4.5~V analog output). For high-pressure characterization up to 1000~kPa, we employed a P-7100-103G-M5 gauge sensor (0.5--4.5~V analog output). We supplied both sensors at 5~V and calibrated them according to manufacturer specifications.

We conducted dynamic blocking tests up to 320 kPa. We performed static sealing tests at 300 kPa for 20 min and during stepwise pressure increases from 300 to 500 kPa. We extracted switching dynamics from time-resolved pressure traces.

\subsection*{Data Analysis}

We recorded pressure signals digitally and converted them from voltage to pressure using the manufacturer-specified linear calibration. Dynamic blocking pressure was defined as the maximum inlet pressure at which the valve could achieve a stable closed state during actuation without detectable downstream pressurization. Blocking thresholds and switching times were extracted directly from the time-resolved pressure traces. For the dynamic blocking-pressure measurements in Fig. 3C(i), each S-EPM condition was measured three times at each pulse voltage, while the Alnico-only valve was measured five times. Data are reported as mean $\pm$ SD. For static blocking measurements, the onset of leakage was defined as an outlet pressure ($P_{out}$) exceeding 3 kPa.

\subsection*{Programmable Fluidic Control and System-Level Demonstrations}

Modular S-EPM valves were assembled into multiple configurations, including dual-tree, circular six-port, decoder, and mix-decoder architectures. Liquid (food dye) and air were used as working media under a constant driving pressure. Valve states ($+1$, $-1$) were applied according to predefined control sequences to achieve selective distribution, combinatorial mixing, and sequential addressing of output channels.

\clearpage

\bibliography{science-ref}
\bibliographystyle{sciencemag}

\section*{Acknowledgments}
\paragraph*{Funding:}
This research was funded by EPSRC Centre for Doctoral Training in Robotics and Autonomous Systems (CDT-RAS), grant number EP/S023208/1.
\paragraph*{Author contributions:}
B.W, J.M, A.A.S: Conceptualization, B.W, J.M: Methodology, Manufacturing, Design, Results, B.W, J.M, F.G.-S, A.A.S: Manuscript drafting.
\paragraph*{Competing interests:}
The authors declare no competing interest.
\paragraph*{Data and materials availability:}
All data supporting the findings of this study are available in the main text and/or supplementary materials.

\end{document}